\documentclass[runningheads]{llncs}
\usepackage{graphicx}
\usepackage{color}

\begin{document}

\title{Attesting Biases and Discrimination using Language Semantics\thanks{Author's copy of the manuscript accepted in the Responsible Artificial Intelligence Agents workshop of the International Conference on Autonomous Agents and Multiagent Systems (AAMAS'19).}}

\author{Xavier Ferrer\inst{1} \and
Jose M. Such\inst{1} \and Natalia Criado\inst{1}}

\institute{Department of Informatics\\
King's College London\\
United Kingdom\\
\email{\{xavier.ferrer\_aran,jose.such,natalia.criado\}@kcl.ac.uk}}

\maketitle            

\begin{abstract}
AI agents are increasingly deployed and used to make automated decisions that affect our lives on a daily basis. It is imperative to ensure that these systems embed ethical principles and respect human values. We focus on how we can attest whether AI agents treat users fairly without discriminating against particular individuals or groups through biases in language. In particular, we discuss human unconscious biases, how they are embedded in language, and how AI systems inherit those biases by learning from and processing human language. Then, we outline a roadmap for future research to better understand and attest problematic AI biases derived from language. 

\end{abstract}

\section{Introduction}\label{sec:intro}

Artificial Intelligence is revolutionising every single aspect of our daily lives: health, banking, insurance, crime prevention, disaster response, social life, culture, and so on and so forth.  AI  is increasingly everywhere: whether embedded in systems or embodied in artefacts such as robots. It affects everyone, and it is transforming public and private organisations and the services and products they offer. While there are no doubts on the benefits of embracing AI technologies, one of the main problems nowadays is ensuring that AI-based systems are ethical and respect human values. 

We particularly focus on digital discrimination caused by AI, which is a form of discrimination in which automated decisions taken by intelligent agents and/or other AI-based systems treat users unfairly, unethically or just differently based on their personal data such as income, education, gender, age, ethnicity or religion \cite{such2017privacy,CriadoDigital2019}.
Digital discrimination is becoming a huge problem \cite{o2016weapons}, as more and more tasks are delegated to AI-based systems, intelligent agents, autonomous systems, and many other systems which embed some kind of AI. For instance, some UK firms base their hiring decisions on AI\footnote{http://www.bbc.co.uk/news/business-36129046}. Examples of digital discrimination include the exhibit of strong female/male gender stereotypes in machine learning systems \cite{bolukbasi2016man}, and the suggestive of arrest records more often shown with online searches of black-sounding names than white-sounding names (regardless of the existence of arrest records
for those names) \cite{sweeney2013discrimination}.

The literature in the area of digital discrimination\footnote{Note the term \textit{digital discrimination} has also been used to define traditional discrimination practices facilitated by online information \cite{edelman2014digital} or a discriminatory access to digital technologies or information \cite{weidmann2016digital}, which we are not using here.} very often refers to the related concept of \textit{algorithmic bias} \cite{danks2017algorithmic}. Despite playing an important role in digital discrimination, bias does not necessarily lead to digital discrimination per se, and this distinction between bias and actual discrimination is crucial. First, a bias usually means a deviation from the standard, but it does not necessarily entail a disadvantageous treatment to particular social groups. For example, an autonomous car that is biased towards safe driving decisions may deviate from the standard driving norms, but it is not  discriminating users. Second, most AI techniques and methods use/need some sort of extent or notion of bias. For instance, machine learning relies on the existence of some statistical patterns in the data used to train them, so that it can learn to predict or make the most suitable decision. Therefore, while bias is a very useful concept and it is indeed related to digital discrimination, we focus on the \textit{problematic} instances of bias, and in the extent to which bias may lead to discrimination.

As interactions between humans and AI agents increasingly happen through language --- e.g., voice-based personal assistants like Amazon Alexa and Google Assistant, and many other AI applications incorporate some kind of natural language processing, the question of problematic instances of bias being learned by AI systems from human language is gaining importance. In this position paper, we discuss human biases, how they are embedded in language, and how AI systems inherit those biases by learning from and processing human language. Then, we outline a roadmap for future research to better understand and assess biases found in language.

\section{Discrimination and Human Biases}
\label{sec:dhlb}
In general, discrimination is related to the notion of fairness, which is usually defined as the impartial and just treatment or behaviour without favouritism. In other words, fairness is based on the principle that any two individuals who are similar with respect to a particular task should be treated with similar benefits and obligations~\cite{dwork2012fairness}. However, the definition of fairness is a concept very much shaped by culture, and closely related to the set of personal and social biases and norms that rule a society: what makes two individuals similar? which set of features are most important when assessing individuals'  similarity?
For instance, skin colour was considered a more important attribute in the United States than in Hong Kong \cite{morland1969race}. On the other hand, in Belgium, Wales, and France, language was deemed to be a salient and relevant characteristic \cite{giles1975speech}. Thus individual (and societal) biases play an important role in determining the definition of fairness, and consequently discrimination. 
Therefore, in order to better understand the various dimensions of discrimination it is key to attest people's implicit and unconscious biases and the social and ethical norms of a society.

From a social psychology perspective, there are methods to determine biases between two concepts (namely unconscious associations), such as the Implicit-Association Test (IAT) \cite{greenwald1998measuring}. 
IAT is a measure designed to detect the strength of a person's automatic association between mental representations of concepts in memory. IAT demonstrates enormous differences in response times when subjects are asked to pair two concepts they find similar, in contrast to two concepts they find different. For instance, a common IAT procedure to detect gender implicit biases consists in the request computer key presses to classify words from four categories, such as female names, male names, words for leader (such as director or chief), and words for helper (such as assistant or employee). When the categories of male and leader are assigned to one key and female and helper to a second,
users usually answer faster than when female and leader share one key (and male and helper share the other). This speed difference indicates the presence of an implicit gender stereotype that associates male with leadership more frequently than female \cite{white2006implicit}.
IAT has been also proved useful to measure racial and ethnic biases~\cite{nosek2007pervasiveness}, religious implicit biases towards homosexuality \cite{rowatt2006associations}, and implicit prejudices based on religious ethnicity, age and nationality \cite{rudman1999measuring}. Therefore, IAT is a useful tool to better understand the behaviours and unconscious biases of individuals.
Unfortunately, IAT is a costly and slow process: one needs first to prepare the test, select a set of subjects and then analyse the results to assess if there exists any implicit bias. Moreover, although IAT might be useful to uncover  biases, it is only applicable when in interaction with human subjects. Consequently, IAT may not be a useful tool to attest biases in AI as such.

\section{Bias in Language}\label{sec:btcs}
Language is used to maintain and transfer culture and cultural ties. Different ideas stem from different uses of language, and can only be fully comprehended when considering the social and cultural context in which they were devised.  Language and culture are closely related \cite{stubbs1996text}: language can be viewed as a verbal expression of culture, and it is known to embed implicit social biases, cultural stereotypes and prejudices \cite{stubbs2001words}.
AI techniques like Natural Language Processing (NLP) have actually been used to extract insights and human unconscious biases from text data \cite{olson1977utterance,hovy2016social}. 
For instance, Fast et.al. \cite{fast2016shirtless} found that modern amateur writers (male and female) are still reinforcing traditional gender stereotypes in their work, by picturing female characters as submissive and male characters as dominant. In \cite{pennebaker2003words}, Pennebacker et.al. found that there exist a strong correlation between age and the use of positive affect words and future tenses: older individuals use more positive affect words.
Regression models
\cite{gilbert2012phrases} and significance tests between the means of lexical categories \cite{kramer2014experimental} are other NLP techniques explored in the literature 
to capture common human stereotypes. 

Given the existence of biases in human language, one could easily hypothesise that AI agents learning from human language would inherit those biases. In fact, Caliskan et.al. \cite{Caliskan2017} proved 
that standard machine learning techniques can acquire stereotypes biases from textual data that reflect everyday human culture, and that text corpora captures semantics including cultural stereotypes and empirical associations. 
To perform their experiments, 
they used \emph{GloVe} \emph{word embeddings} \cite{pennington2014glove} with a general corpus of text from the Web.\footnote{They used a GloVe word embedding model trained from a 840 billion word corpus} 
Word embeddings~\cite{goldberg2016primer} are vector word representations that encode the textual context in which a word is found in a high-dimensional space. The process of learning word embeddings involves associating words with parameter vectors that are optimised to be good predictors of other words occurring in similar contexts. For this reason, word embedding representations can be used to estimate the association between words: two words with similar embeddings can be considered to be semantically related.
Based on word embedings, Caliskan et al. \cite{Caliskan2017} defined what they called the Word-Embedding Association Test (WEAT), as the cosine similarity between pairs of word embedding vectors. Using WEAT they showed that the distance between the embeddings of two words is analogous to reaction time in the IAT by replicating the results obtained in some well-known psychological studies related with IAT. They were able to replicate
stereotypes tested in IATs that studied general societal attitudes for which lists of target and attribute words were available,
including general societal discriminative attitudes observed in the experiments performed by Greenwald and Bertrand et.al.  \cite{greenwald1998measuring,bertrand2004emily}.
Among other findings, and according to the corresponding IAT experiments, they found that in the word embeddings vector space European American names were more likely than African American names to be closer to pleasant than to unpleasant bearing words, and that female names were associated more frequently to family words instead of career words than male names. Caliskan et. al. were also able to correctly predict the percentage of women in 50 occupations as presented in the 2015 U.S. Bureau of Labour Statistics by analysing the word embedding distance of a set of female-sounding names to the corresponding occupations.

Another relevant work using word embeddings is described in \cite{bolukbasi2016man}. The authors found that a word2vec\footnote{https://code.google.com/archive/p/word2vec/
} word embedding model trained with a public corpus of Google News articles exhibits strong female/male gender stereotypes. Moreover, they described a method for removing gender bias from word embeddings by comparing and adapting the differences between the direction of the gender-biased word embedding vectors with those gender-neutral vectors. This debiasing procedure is criticised in \cite{Caliskan2017}, arguing that the process of debiasing alters the AI’s model of the world instead of how the AI acts on that perception -- which can be considered a case of \emph{fairness through blindness \cite{dwork2012fairness}}.

\section{Towards Attesting Bias and Discrimination in Language}
We have seen than written language carries the personal and implicit biases of its authors, and that these biases can be later learned by AI models trained with this data. 
In addition, the automated decisions made by these models help perpetuating the (same) stereotypes and biases embedded in the language corpora used for training.
A clear example of this process is seen in  \cite{Caliskan2017,bolukbasi2016man}, as
the dataset they used was formed by a collection of news from a specific country and cultural context, and by feeding this data into a machine learning system (a word embeddings model) they were able to reproduce unconscious human biases similar to those discovered with IAT tests obtained from individuals with similar cultural contexts. The deployment of these word embedding models in a real word environment leads to, as shown in \cite{Caliskan2017,bolukbasi2016man}, perpetuating the same implicit human biases present in the news corpora used to train the word embeddings model. 
Therefore, we believe that working towards a general framework to attest the nature of digital discrimination and the role that biases embedded in text corpora play in it is of crucial importance to develop ethical AI. We particularly envision two main lines of work in this regard, which we describe in detail below: i)
advancing in the discovery and understanding of human biases and discrimination embedded in text corpora generated by humans, and ii) attesting biases and discrimination given any language corpus -- including corpora generated by AI agents and systems with embedded AI.

\paragraph{\textbf{Understanding human biases.}}
A very important step towards a general framework to attest digital discrimination in AI is further understanding human biases and discrimination, since they are human constructs closely related with culture and social norms. 
For instance, since the cultural and social dimensions of individuals play an important role in defining their implicit biases \cite{giles1975speech,morland1969race}, it seems clear that people from similar cultural contexts may share similar implicit biases. Can we identify which biases are related to specific society's cultural and ethical norms? Can we identify similar cultures and social values by analysing the implicit biases incorporated in text corpora authored by members of a society?

On the other hand, closely related to human biases is the well studied theory of \emph{social identity} \cite{tajfel1979integrative}. Social identity theory states that the person's sense of herself is based on the groups to which an individual belongs. It is believed that we choose social groups based on the stereotypes attached to each social identity, and that these selections impact our behaviour with other individuals and groups. On top of that, the \emph{affect control theory} \cite{heise1979understanding} provides a predictive model for how stereotypes are used to label social identities, groups, and relations \cite{heise1987affect}. 
Therefore, we might be able to study how the different ethnic groups and prejudices are related by analysing text corpora semantics. Is the concept of social identities and affect control theory encoded in a corpus of texts by means of the implicit biases of the author's? Can we identify society's discriminated groups by analysing language? Are 
the affective stereotypes between different social identities \cite{heise2010surveying} stable across time?

Finally, a very exciting line of research would be to analyse whether word embedding models or other NLP techniques embed the particular social biases that led to discrimination found in real discrimination cases. We have seen that word embedding models contain gender biases and strong female/male gender stereotypes \cite{bolukbasi2016man,Caliskan2017}.
However, do word embeddings encode the social biases (e.g. a closer semantic association between specific ethnic groups to criminality) that led to the automated decision of showing more often arrest records for black-sounding names than for white-sounding names in online searches \cite{sweeney2013discrimination}? 

\paragraph{\textbf{Understanding and attesting AI biases.}}

In parallel and building on the advances above, another interesting line of research would be to explore AI biases by analysing the texts authored or speech synthesised by AI agents. Can we discover, given a language corpus created by an AI agent (e.g. a chat log created by interacting with a chat bot), the biases the AI agent may have? To which extent are human and AI biases related? It may seem that, since most AI agents are trained with human-authored corpora, there should be traces of human-biases learned by AIs and therefore present in AI-generated text corpora. However, machine learning systems are able to learn and reason by \emph{themselves}, sometimes in ways not completely transparent to humans. Therefore, will AI leave imprints of their own biases when authoring text-based corpora?

Suppose now a black box scenario. That is, a scenario in which an individual has control over the inputs and outputs of an AI agent, but cannot access the inner workings or the data used to train the AI agent. Can biases encoded in text corpora authored by an AI agent give insights about the data used to train the AI agent? If so, by leveraging those insights, can we discover how the black box operates? For instance, consider the case of smart personal assistants (such as Alexa or Google Home). Both of them are black boxes: it is unknown, except for Amazon and Google, the datasets used to train the model in charge of translating speech-to-text. 
Would it be possible, by interacting long enough with the smart assistant, to discover hidden biases in the usage of language used by the assistants? Can we determine the social and cultural environment of the authors of the datasets used to train them by leveraging the information encoded in the semantics? Knowing the dataset used to train those black box systems can be an important step towards discovering how these systems work, and to assess whether their unconscious biases may lead to discriminatory outcomes.

It would also be interesting to analyse how bias and discrimination are gradually learnt and incorporated into the language of a reinforcement learning system when in interaction with biased opinions. 
For instance, consider the case of Microsoft Twitter chatbot \emph{Tay},
a bot that responded to users queries emulating the casual speech
patterns of a stereotypical millennial while learning from his Twitter conversations with people. The experiment ended in less than 24 hours,
when Microsoft decided to shut down \emph{Tay} after it started tweeting racist and discriminatory slogans learned in tweets \cite{price2016microsoft}. 

Finally, if we can attest biases in an AI agent looking at the language it uses, could not the AI agent itself realise about this? What kind of mechanisms can we develop so as to give the AI agent the power to, understanding and realising about its own learned biases, address them or consider them to make fairer decisions? Of particular interest here would be the combination of data-driven approaches like machine learning and knowledge-based approaches like normative systems as a way forward for making AI agents aware of where learned biases would not be acceptable.

\section{Conclusions}

Since AI agents and systems are increasingly deployed and used to make automated decisions that affect our lives on a daily basis, it is imperative to ensure that these systems embed ethical principles and respect human values. In particular, we focus on how we can attest whether AI agents treat users fairly without discriminating against particular individuals or groups. 
We propose to advance in this line of research by: i) exploring the usage of language, a cultural vehicle that embeds the prejudices and the implicit ethical  biases of a society, and ii) exploring human-like biases and discrimination embedded in language corpora and corpora generated by AI agents and systems with embedded AI. This would be the basis for a general framework to attest digital discrimination by AI and the role that biases embedded in language corpora play in it.

\subsubsection{Acknowledgments.}
We would like to thank the EPSRC for supporting this research under grant DADD: Discovering and Attesting Digital Discrimination (EP/R033188/1), and NVIDIA Corporation for the donation of the Titan Xp.

\bibliographystyle{abbrv}
\bibliography{bib}

\begin{thebibliography}{10}

\bibitem{bertrand2004emily}
M.~Bertrand and S.~Mullainathan.
\newblock Are emily and greg more employable than lakisha and jamal? a field
  experiment on labor market discrimination.
\newblock {\em American economic review}, 94(4):991--1013, 2004.

\bibitem{bolukbasi2016man}
T.~Bolukbasi, K.-W. Chang, J.~Y. Zou, V.~Saligrama, and A.~T. Kalai.
\newblock Man is to computer programmer as woman is to homemaker? debiasing
  word embeddings.
\newblock In {\em Advances in Neural Information Processing Systems}, pages
  4349--4357, 2016.

\bibitem{Caliskan2017}
A.~Caliskan, J.~J. Bryson, and A.~Narayanan.
\newblock {Supplementary Materials for: Semantics derived automatically from
  language corpora contain human-like biases}.
\newblock {\em Science}, 356(6334):183--186, 2017.

\bibitem{CriadoDigital2019}
N.~Criado and J.~M. Such.
\newblock {Digital Discrimination}.
\newblock In {\em Algorithmic Regulation}, chapter Digital Discrimination.
  Oxford University Press, 2019.

\bibitem{danks2017algorithmic}
D.~Danks and A.~J. London.
\newblock Algorithmic bias in autonomous systems.
\newblock In {\em Proc. of the Int. Joint Conf. on Artificial Intelligence},
  pages 4691--4697. AAAI Press, 2017.

\bibitem{dwork2012fairness}
C.~Dwork, M.~Hardt, T.~Pitassi, O.~Reingold, and R.~Zemel.
\newblock Fairness through awareness.
\newblock In {\em Proceedings of the 3rd innovations in theoretical computer
  science conference}, pages 214--226. ACM, 2012.

\bibitem{edelman2014digital}
B.~G. Edelman and M.~Luca.
\newblock Digital discrimination: The case of airbnb.
\newblock 2014.

\bibitem{fast2016shirtless}
E.~Fast, T.~Vachovsky, and M.~S. Bernstein.
\newblock Shirtless and dangerous: Quantifying linguistic signals of gender
  bias in an online fiction writing community.
\newblock In {\em ICWSM}, pages 112--120, 2016.

\bibitem{gilbert2012phrases}
E.~Gilbert.
\newblock Phrases that signal workplace hierarchy.
\newblock In {\em Proc. of the ACM 2012 conference on Computer Supported
  Cooperative Work}, pages 1037--1046. ACM, 2012.

\bibitem{giles1975speech}
H.~Giles and P.~Powesland.
\newblock {\em Speech style and social evaluation.}
\newblock Ac. Press, 1975.

\bibitem{goldberg2016primer}
Y.~Goldberg.
\newblock A primer on neural network models for natural language processing.
\newblock {\em Journal of Artificial Intelligence Research}, 57:345--420, 2016.

\bibitem{greenwald1998measuring}
A.~G. Greenwald, D.~E. McGhee, and J.~L. Schwartz.
\newblock Measuring individual differences in implicit cognition: the implicit
  association test.
\newblock {\em Journal of personality and social psychology}, 74(6):1464, 1998.

\bibitem{heise1979understanding}
D.~R. Heise.
\newblock {\em Understanding events: Affect and the construction of social
  action}.
\newblock CUP Archive, 1979.

\bibitem{heise1987affect}
D.~R. Heise.
\newblock Affect control theory: Concepts and model.
\newblock {\em Journal of Mathematical Sociology}, 13(1-2):1--33, 1987.

\bibitem{heise2010surveying}
D.~R. Heise.
\newblock {\em Surveying cultures: Discovering shared conceptions and
  sentiments}.
\newblock John Wiley \& Sons, 2010.

\bibitem{hovy2016social}
D.~Hovy and S.~L. Spruit.
\newblock The social impact of natural language processing.
\newblock In {\em Proc. of Association for Computational Linguistics},
  volume~2, pages 591--598, 2016.

\bibitem{kramer2014experimental}
A.~D. Kramer, J.~E. Guillory, and J.~T. Hancock.
\newblock Experimental evidence of massive-scale emotional contagion through
  social networks.
\newblock {\em Proceedings of the National Academy of Sciences}, page
  201320040, 2014.

\bibitem{morland1969race}
J.~K. Morland.
\newblock Race awareness among american and hong kong chinese children.
\newblock {\em American Journal of Sociology}, 75(3):360--374, 1969.

\bibitem{nosek2007pervasiveness}
B.~A. Nosek, F.~L. Smyth, J.~J. Hansen, T.~Devos, N.~M. Lindner, K.~A.
  Ranganath, C.~T. Smith, K.~R. Olson, D.~Chugh, A.~G. Greenwald, et~al.
\newblock Pervasiveness and correlates of implicit attitudes and stereotypes.
\newblock {\em European Review of Social Psychology}, 18(1):36--88, 2007.

\bibitem{olson1977utterance}
D.~Olson.
\newblock From utterance to text: The bias of language in speech and writing.
\newblock {\em Harvard educational review}, 47(3):257--281, 1977.

\bibitem{o2016weapons}
C.~O'Neil.
\newblock {\em Weapons of math destruction: How big data increases inequality
  and threatens democracy}.
\newblock Broadway Books, 2016.

\bibitem{pennebaker2003words}
J.~W. Pennebaker and L.~D. Stone.
\newblock Words of wisdom: Language use over the life span.
\newblock {\em Journal of personality and social psychology}, 85(2):291, 2003.

\bibitem{pennington2014glove}
J.~Pennington, R.~Socher, and C.~Manning.
\newblock Glove: Global vectors for word representation.
\newblock In {\em EMNLP Proceedings}, pages 1532--1543, 2014.

\bibitem{price2016microsoft}
R.~Price.
\newblock Microsoft is deleting its ai chatbot’s incredibly racist tweets.
\newblock {\em Business Insider}, 2016.

\bibitem{rowatt2006associations}
W.~C. Rowatt, J.-A. Tsang, J.~Kelly, B.~LaMartina, M.~McCullers, and
  A.~McKinley.
\newblock Associations between religious personality dimensions and implicit
  homosexual prejudice.
\newblock {\em Journal for the Scientific Study of Religion}, 45(3):397--406,
  2006.

\bibitem{rudman1999measuring}
L.~A. Rudman, A.~G. Greenwald, D.~S. Mellott, and J.~L. Schwartz.
\newblock Measuring the automatic components of prejudice: Flexibility and
  generality of the implicit association test.
\newblock {\em Social cognition}, 17(4):437--465, 1999.

\bibitem{stubbs1996text}
M.~Stubbs.
\newblock {\em Text and corpus analysis: Computer-assisted studies of language
  and culture}.
\newblock Blackwell Oxford, 1996.

\bibitem{stubbs2001words}
M.~Stubbs.
\newblock {\em Words and phrases: Corpus studies of lexical semantics}.
\newblock Blackwell Publishers Oxford, 2001.

\bibitem{such2017privacy}
J.~M. Such.
\newblock Privacy and autonomous systems.
\newblock In {\em Proc. of the 26th International Joint Conference on
  Artificial Intelligence}, pages 4761--4767. AAAI Press, 2017.

\bibitem{sweeney2013discrimination}
L.~Sweeney.
\newblock Discrimination in online ad delivery.
\newblock {\em Queue}, 11(3):10, 2013.

\bibitem{tajfel1979integrative}
H.~Tajfel, J.~C. Turner, W.~G. Austin, and S.~Worchel.
\newblock An integrative theory of intergroup conflict.
\newblock {\em Organizational identity: A reader}, pages 56--65, 1979.

\bibitem{weidmann2016digital}
N.~B. Weidmann, S.~Benitez-Baleato, P.~Hunziker, E.~Glatz, and
  X.~Dimitropoulos.
\newblock Digital discrimination: Political bias in internet service provision
  across ethnic groups.
\newblock {\em Science}, 353(6304):1151--1155, 2016.

\bibitem{white2006implicit}
M.~J. White and G.~B. White.
\newblock Implicit and explicit occupational gender stereotypes.
\newblock {\em Sex roles}, 55(3-4):259--266, 2006.

\end{thebibliography}

\end{document}